\documentclass[wcp]{jmlr}

\usepackage{longtable}% for long tables

\usepackage{booktabs}
\usepackage{graphicx} % more modern

\usepackage{algorithm}
\usepackage{algorithmic}
\usepackage{multirow}
\usepackage{enumitem,calc}

\newtheorem{pro}{Proposition}

\jmlrvolume{60}
\jmlryear{2016}
\jmlrworkshop{ACML 2016}

\title[Short Title]{Selecting Bases in Spectral learning of Predictive State Representations via Model Entropy}

  \author{\Name{Yunlong Liu} \Email{ylliu@xmu.edu.cn}\and
   \Name{Hexing Zhu} \Email{hxzhu$\_$xmu@foxmail.com}\\
   \addr Department of Automation, Xiamen University, Xiamen, China, 361005}

\begin{document}

\maketitle

\begin{abstract}
Predictive State Representations~(PSRs) are powerful techniques for modelling dynamical systems, which represent a state as a vector of predictions about future observable events~(tests). In PSRs, one of the fundamental problems is the learning of the PSR model of the underlying system. Recently, spectral methods have been successfully used to address this issue by treating the learning problem as the task of computing an singular value decomposition~(SVD) over a submatrix of a special type of matrix called the Hankel matrix. Under the assumptions that the rows and columns of the submatrix of the Hankel Matrix are sufficient~(which usually means a very large number of rows and columns, and almost fails in practice) and the entries of the matrix can be estimated accurately, it has been proven that the spectral approach for learning PSRs is statistically consistent and the learned parameters can converge to the true parameters. However, in practice, due to the limit of the computation ability, only a finite set of rows or columns can be chosen to be used for the spectral learning. While different sets of columns usually lead to variant accuracy of the learned model, in this paper, we propose an approach for selecting the set of columns, namely basis selection, by adopting a concept of model entropy to measure the accuracy of the learned model. Experimental results are shown to demonstrate the effectiveness of the proposed approach.
\end{abstract}

\section{Introduction}

Modelling dynamical systems is a common problem in science and engineer, and there are many applications related to such a modelling, e.g. robot navigation, natural language processing, speech recognition, etc.~\cite{ICML2013_hamilton13}. Many approaches have been devoted to modelling systems, such as Hidden Markov Models~(HMM) for uncontrolled dynamical systems and Partially Observable Markov Decision Processes~(POMDPs) for controlled systems. However, such latent-state approaches usually suffer from local minima and require some assumptions~\cite{Singh04predictivestate}. Predictive State Representations~(PSRs) offer an effective approach for modelling partially dynamical systems~\cite{Littman01predictiverepresentations}. Unlike the latent-state approaches, PSRs use a vector of predictions about future events, called tests, to represent the state. The tests can be executed on the systems and are fully observable. Compared to the latent-state approaches, PSRs have shown many advantages, such as the possibility of obtaining a global optimal model, more expressive power and less required prior domain knowledge, etc.~\cite{LiuAAMAS15}.

There are two main problems in PSRs. One is the learning of the PSR model; the other is the application of the learned model, including predicting and planning~\cite{Rosencrantz04learninglow,Liu14inaccuratePSR}. The state-of-the-art technique for addressing the learning problem is the spectral approach~\cite{Boots-2011b}. Spectral methods treat the learning problem as the task of computing a singular value decomposition~(SVD) over a submatrix $H_s$ of a special type of matrix called the Hankel matrix~\cite{Hsu_aspectral}. Under the strong assumptions that the rows $H$ and columns $T$ of $H_s$ are sufficient and the entries of the matrix can be estimated accurately, it has been proven that the spectral approach for learning PSRs is statistically consistent and the learned parameters can converge to the true parameters~\cite{Boots-2011b}. However, for the sufficient assumption, it usually means a very large number of rows or columns of the $H_s$ matrix, which almost fails in practice~\cite{kulesza2015spectral}. At the same time, the computation complexity of the SVD operation on $H_s$ is $O(|T|^2|H|)$~($|T|$ and $|H|$ is the number of the columns and rows in the matrix respectively), for large set of $T$ and $H$, such an operation is also prohibitively expensive. Also, for the spectral approach, to obtain the model parameter,  one should estimate and manipulate two observable matrices $P_{T,H}$, $P_{H}$, and $|A|\times|O|$ observable matrices $P_{T,ao,H}$~\cite{Liu16IJCAI}. Although Denis {\em et al.}~\cite{DBLP:conf/icml/DenisGH14} showed that the concentration of the empirical $P_{T,H}$ around its mean does not highly depend on its dimension, which gives a hope for alleviating the statistical problem when using large set of $T$ and $H$ as the accuracy of the learned model is directly connected to the concentration, manipulating these matrices is still too expensive to afford in large systems.

Thus, in practice, taking the computational constraints into account, it is needed to find only a finite set of columns of the Hankel matrix before the spectral methods can be applied. While different sets of columns usually lead to variant accuracy of the learned model, in this paper, we first introduce a concept of model entropy and show the high relevance between the entropy and the accuracy of the learned model, then for the columns selection problem, we call it basis selection in spectral learning of PSRs, an approach is proposed by using the model entropy as the guidance. We finally show the effectiveness of the proposed approach by executing experiments on {\em PocMan}, which is one benchmark domain in the literature.

We organize the remainder of this paper as follows. We briefly review the background and define notations in Section 2. We propose the approaches for basis selection in Section 3. We provide comparative results in Section 4 . Finally we conclude the paper in Section 5.

\section{Preliminaries}

Predictive state representations~(PSRs) represent state by using a vector of predictions of fully observable quantities~(tests) conditioned on past events~(histories), denoted $b(\cdot)$. For discrete systems with finite set of observations $O=\{o^1,o^2,\cdots,o^{|O|}\}$ and actions $A=\{a^1,a^2,\cdots,a^{|A|}\}$, at time $\tau$, a \emph{test} is a sequence of action-observation pairs that starts from time $\tau + 1$. Similarly, a \emph{history} at $\tau$ is a sequence of action-observation pairs that starts from the beginning of time and ends at time $\tau$, which is used to describe the full sequence of past events. The prediction of a length-$m$ test $t$ at history $h$ is defined as $p(t|h)=p(ht)/p(h)=\prod^m_{i=1}Pr(o_i|ha_1o_1\cdots a_i)$~\cite{Singh04predictivestate}.

The underlying dynamical system can be described by a special bi-infinite matrix, called the Hankel matrix~\cite{Balle:2014ML}, where the rows and columns correspond to all the possible tests $T$ and histories $H$ respectively, the entries of the matrix are defined as $P_{t,h}=p(ht)$ for any $t \in T$ and $h \in H$, where $ht$ is the concatenation of $h$ and $t$~\cite{Boots-2011b}. The rank of the Hankel matrix is called the linear dimension of the system. When the rank is finite, we assume it is $k$, in PSRs, the state of the system at history $h$ can be represented as a prediction vector of $k$ tests conditioned at $h$. The $k$ tests used as the state representation is called the minimal {\em core tests} that the predictions of these tests contain sufficient information to calculate the predictions for all tests, and is a \emph{sufficient statistic}~\cite{Singh04predictivestate,LiuAAMAS15}. For linear dynamical systems, the minimal core tests can be the set of tests that corresponds to the $k$ linearly independent columns of the Hankel matrix~~\cite{LiuAAMAS15}.

A PSR of rank $k$ can be parameterized by a reference condition state vector $b_* = b(\epsilon)\in \mathbb{R}^k$, an update matrix $B_{ao} \in \mathbb{R}^{k \times k}$ for each $a \in A$ and $o \in O$, and a normalization vector $b_\infty \in \mathbb{R}^k$, where $\epsilon$ is the empty history and $b_\infty^T B_{ao} = 1^T$~\cite{Hsu_aspectral,Boots-2011b}. In the spectral approach, these parameters can be defined in terms of the matrices $P_\mathcal{H}$, $P_{\mathcal{T,H}}$, $P_{\mathcal{T},ao,\mathcal{H}}$ and an additional matrix $U \in \mathbb{R}^{T \times d}$ as shown in Eq.~\ref{equ:spectral}, where $\mathcal{T}$ and $\mathcal{H}$ are the set of all possible tests and histories respectively, $U$ is the left singular vectors of the matrix $P_{\mathcal{T,H}}$, $^T$ is the transpose and $\dag$ is the pseudo-inverse of the matrix~\cite{Boots-2011b}.
\begin{equation}
\begin{split}
& b_* = U^TP_{\mathcal{H}}1_k, \\
& b_{\infty} = (P^T_{\mathcal{T,H}})^{\dag}P_\mathcal{H}, \\
& B_{ao} = U^TP_{\mathcal{T},ao,\mathcal{H}}(U^TP_{\mathcal{T,H}})^\dag.
\end{split}
\label{equ:spectral}
\end{equation}

Using these parameters, after taking action $a$ and seeing observation $o$ at history $h$, the PSR state at next time step $b(hao)$ is updated from $b(h)$ as follows~\cite{Boots-2011b}:

\begin{equation}
b(hao) = \frac{B_{ao}b(h)}{b_\infty^TB_{ao}b(h)}.
\end{equation}

Also, the probability of observing the sequence $a_1o_1a_2o_2 \cdots a_no_n$ in the next $n$ time steps can be  predicted by~\cite{kulesza2015spectral}:

\begin{equation}
Pr[o_{1:t}||a_{1:t}] = b_{\infty}^TB_{a_no_n} \cdots B_{a_2o_2}B_{a_1o_1}b_*.
\end{equation}

Under the assumption that the columns and rows of the submatrix $H_s$ of the Hankel matrix are sufficient, when more and more data are included, the law of large numbers guarantees that the estimates $\hat P_\mathcal{H}$, $\hat P_\mathcal{T,H}$, and $\hat P_{\mathcal{T},ao,\mathcal{H}}$ converge to the true matrices $P_\mathcal{H}$, $P_\mathcal{T,H}$, and $P_{\mathcal{T},ao,\mathcal{H}}$, the estimates $\hat b_*$, $\hat b_{\infty}$, and $\hat B_{ao}$ converge to the true parameters $b_*$, $b_{\infty}$, and $B_{ao}$ for each $a \in A$ and $o \in O$, that the learning is consistent~\cite{Boots-2011b}.

\section{Basis Selection via Model Entropy}
As the assumption that the columns and rows of the submatrix $H_s$ are sufficient is really strong~(usually means a very large set of columns and histories of the matrix), which almost fails in reality. At the same time, due to the computation and statistics constraints, we can only manipulate a limited finite set of tests/histories. As different sets of columns~(tests) or rows~(histories) usually cause different model accuracy, which requires us to select the finite set of tests and rows~(histories) of the matrix for applying the spectral approach. In practice, as all possible training data is used to generate the histories, how to select the tests, i.e., the basis selection, is the crucial problem. In this section, we first introduce the concept of model entropy, then show the high relevance between the entropy and the model accuracy, finally we propose a simple search method for selecting the bases.

\noindent{\bf Model Entropy}. Given a set of tests $X=\{x_1, x_2, \cdots, x_i\}$, if it includes the set of core tests and the number of possible $p(X|\cdot)$ is finite, Proposition~\ref{pro:mdp} holds~\cite{Liu16IJCAI}.
\begin{pro}
The Markov decision process~(MDP) model built by using $p(X|\cdot)$ as state representation and the action-observation pair $\langle ao \rangle$ as action is deterministic.
\label{pro:mdp}
\end{pro}
In practice, it is difficult for $X$ to include the set of core tests. At any time step, the prediction vector $p(X|\cdot)$ may actually correspond to several PSR states. In such cases, the transition from $p(X|h)$ to $p(X|hao)$ usually becomes stochastic and the less information is included in $X$, the more stochastic the transition will usually be. Inspired by the concept of Shannon entropy that measures information uncertainty~\cite{shannon48}, to quantify the stochasticity, a concept of model entropy for each set of tests $X$ is defined in  Eq.~\ref{equ:entropy}~\cite{Liu16IJCAI}.
\begin{equation}
\begin{small}
%E_M = \sum_{a \in \mathcal{A}_{PP}} \frac{1}{r(T^a)} \sum_{i=1}^{r(T^a)} \sum_{j=1}^{c(T^a)}T(s_i,a,s_j)\log \frac{1}{T(s_i,a,s_j)}
E(X) = -\sum_{a \in \mathcal{A}_{PP}} \frac{1}{r(T^a)} \sum_{i=1}^{r(T^a)} \sum_{j=1}^{c(T^a)}T(s_i,a,s_j)\log T(s_i,a,s_j),
\end{small}
\label{equ:entropy}
\end{equation}
where $T$ is the state-transition function of the MDP using $p(X|\cdot)$ as state representation and $\langle ao \rangle$ as action, $\mathcal{A}_{PP}=A \times O$ the set of action-observation pairs in the original system, $r(T^a)$ and $c(T^a)$  the number of rows and columns of the state-transition matrix $T^a$ respectively.

\noindent{\bf Relevance between Model Entropy and Basis Selection}. When selecting the set of tests for applying spectral methods, the tests containing more information should be selected as the more information included in the set of tests, the higher accuracy the learned PSR model will usually be. According to Eq.~\ref{equ:entropy}, the less information is included in the set of tests $X$, the more stochastic the state transition will be, and the higher the model entropy is. So for the basis selection problem, the set of tests with lowest model entropy should be selected.

\noindent{\bf Basis Selection via Model Entropy}. Using the model entropy as the guidance, we propose a simple local search algorithm for searching the set of tests used for spectral learning of PSRs, which is shown in Algorithm~\ref{Algo:selecting}. Starting with a default $T$ of the desired size, we iteratively sample a set of new tests and consider using it to replace the tests in $T$. If the replacement is a reduction in terms of the entropy value, then we keep it. After a fixed number of rounds, we stop and return the current $T$. In the algorithm, the entropy $E(T)$ for each candidate set of tests $T$ is calculated as follows~(We name this procedure as $EntropyLearn(D,T)$): We first generate an original randomly action-observation sequences $D$ as the training data for calculating the entropy. Then the data is translated into the form of $\langle$action-observation$\rangle$-$p(T|\cdot)$ sequences. For example, a sequence $d=\langle a_1o_1a_2o_2 \cdots a_ko_k \rangle$ is converted into $d'=\langle a_1o_1 \rangle p(T|a_1o_1)\langle a_2o_2\rangle p(T|a_1o_1a_2o_2) \cdots \langle a_ko_k\rangle p(T|a_1o_1a_2o_2 \cdots a_ko_k)$, where $p(T|\cdot)$ can be estimated using the training data. Due to the sampling error, it is unlikely that any of these estimated $p(\hat{T}|\cdot)$ will be exactly the same, even if the true underlying $p(T|\cdot)$ are identical. Statistical tests or linearly independent techniques can be used to estimate the number of distinct underlying $p(T|\cdot)$ and cluster the estimated $p(\hat{T}|\cdot)$ corresponding to the same true prediction vector into one group(state)~\cite{TalvitieS11,Liu14inaccuratePSR}. Subsequently, we compute the state-transition functions in the transformed data and build the MDP model. Finally, the entropy $E(T)$ of a set of tests $T$ can be calculated according to Eq.~\ref{equ:entropy}.

\begin{algorithm}[tb]
   \caption{Search for sets of $k$ tests $T$ that approximately minimize $E(T)$ }
   \label{Algo:selecting}
   %\SetVline
\begin{algorithmic}[1]
   \STATE {\bfseries Input:} dataset $D$, initial $T$ of size $k$, number of size $n < k$, distributions $p_T$ over candidate tests, number of rounds $r$, entropy value $EV$, an entropy threshold $E_{th}>0$, number of iteration $iterNum$.
   \STATE {\bfseries Initialize:} $EV := EntropyLearn(D,T)$.
   \FOR{$i=1$ {\bfseries to} $r$}
   \FOR{$j=1$ {\bfseries to} $iterNum$}
   \STATE {Sample a set of $n$ tests $T_s$ $\not \in T \sim p_T$}
   \STATE {$T^{'}$ $\leftarrow$ a set of $n$ test in $T$}
   \STATE {$EV_T := EntropyLearn(D,T \setminus T^{'} \cup T_s)$}
   \IF{$EV - EV_T > E_{th}$}
   \STATE $EV := EV_T$
   \STATE $T := T \setminus T^{'} \cup T_s$
   \STATE {\bfseries break}
   \ENDIF
   \ENDFOR
   \ENDFOR
   \STATE {\bfseries Output:} $T$
\end{algorithmic}
\end{algorithm}

\section{Experimental Results}
We evaluated the proposed technique on {\em PocMan}~\cite{Silver10monte-carloplanning,JMLR:hamilton14}, a partially observable version of the video game {\em PacMan}. The environment is extremely large that has 4 actions, 1024 observations and an extremely large number of states~(up to $10^{56}$).

\subsection{Experimental Setting}
\noindent{\bf Evaluated Methods}.
We compared our approach~(Entropy-based) with the bound-based approach~\cite{kulesza2015spectral}, which is the most related technique with our method that both the approaches address the same basis selection problem. The two approaches both first select a set of tests, then spectral method is applied by using these tests. Also, iterative approaches are used in both approaches for selecting the sets of tests. The main difference between our approach and the work of~\cite{kulesza2015spectral} is the guidance used for selecting the set of tests. While our approach uses the model entropy as the guidance, the key idea in~\cite{kulesza2015spectral} is that as in the limiting-case, the singular values of the learned transformed predictive state representations~(TPSR) parameters are bounded, then under an assumption that the smaller the singular values of the learned parameters are, the more accuracy the learned PSR model will be, they just select the tests that can cause smaller singular values of the learned PSR parameters. However, no any formal guarantees are provided for such an approach.

A uniformly random generated  sequence with length $200,000$ was used as the training sequence for both approaches. And to calculate the entropy of a candidate set of tests $T_c$, a randomly generated 5,000-length action-observation sequence $d$ was used. For each test $t \in T_c$, $p(t|\cdot)$ was estimated by executing $act(t)$ 100 times, where $act(t)$ is the action sequence in test $t$. The entropy thresholds for our approaches are 0.06, 0.04, and 0.02 for different number of tests as the entropy value usually decreases with the increase of the number of tests~(For number of tests 100 and 150, the threshold used is 0.06; for number of tests 200 and 250, the threshold is 0.04; and for number of tests 300, the threshold is 0.02). The number of rounds for both approaches is 10. In each round, we iteratively sample a set of new tests of size 20 and use it to replace 20 tests in $T$, the iterative number in each round is also 10. \iffalse if the replacement is an reduction in terms of the entropy value or the bound in 10 iterations~(the reduction no less than the threshold for our approach), then we keep it and go to next round, otherwise, we also go to next round.\fi After a fixed number of rounds, we stop and return the current $T$.

\noindent{\bf Performance Measurements.}
We evaluated the learned models in terms of prediction accuracy, which is measured by the difference between the true predictions  and the predictions given by the learned model over a test data sequence~(for {\em PocMan}, we cannot obtain the true predictions, Monte-Carlo rollout predictions  were used~\cite{JMLR:hamilton14}).

Two error functions were used in the measurement. One is the average one-step prediction error per time step on the test sequence as shown in Eq.~\ref{eqn:f1}:

\begin{equation}
\frac{1}{L} \sum_{t=1}^L |p(o_{t+1}|h_t,a_{t+1})-\hat{p}(o_{t+1}|h_t,a_{t+1})|.
\label{eqn:f1}
\end{equation}
$p(\cdot)$ is the probability calculated from the true POMDP model or the Monte-Carlo rollout prediction and $\hat{p}(\cdot)$ the probability obtained from the learned model. $|\cdot|$ refers to the absolute value. $L$ is the length of the test sequence used in the experiments. For the average four-step prediction error, the same equation with predicting four steps ahead $\hat{p}(o_{t+1}o_{t+2}o_{t+3}o_{t+4}|h_ta_{t+1}a_{t+2}a_{t+3}a_{t+4})$ was used.

\subsection{Performance Evaluation}

In this section, for each algorithm, we report the performance results as the mean error over 10 trials. For each trial, a uniform randomly generated test sequence with length $L=20,000$ was used for testing the accuracy of the learned model.

\begin{figure*}[!ht]
\begin{minipage}{6.0in}
\begin{minipage}{3in}
\includegraphics[width=0.8\textwidth]{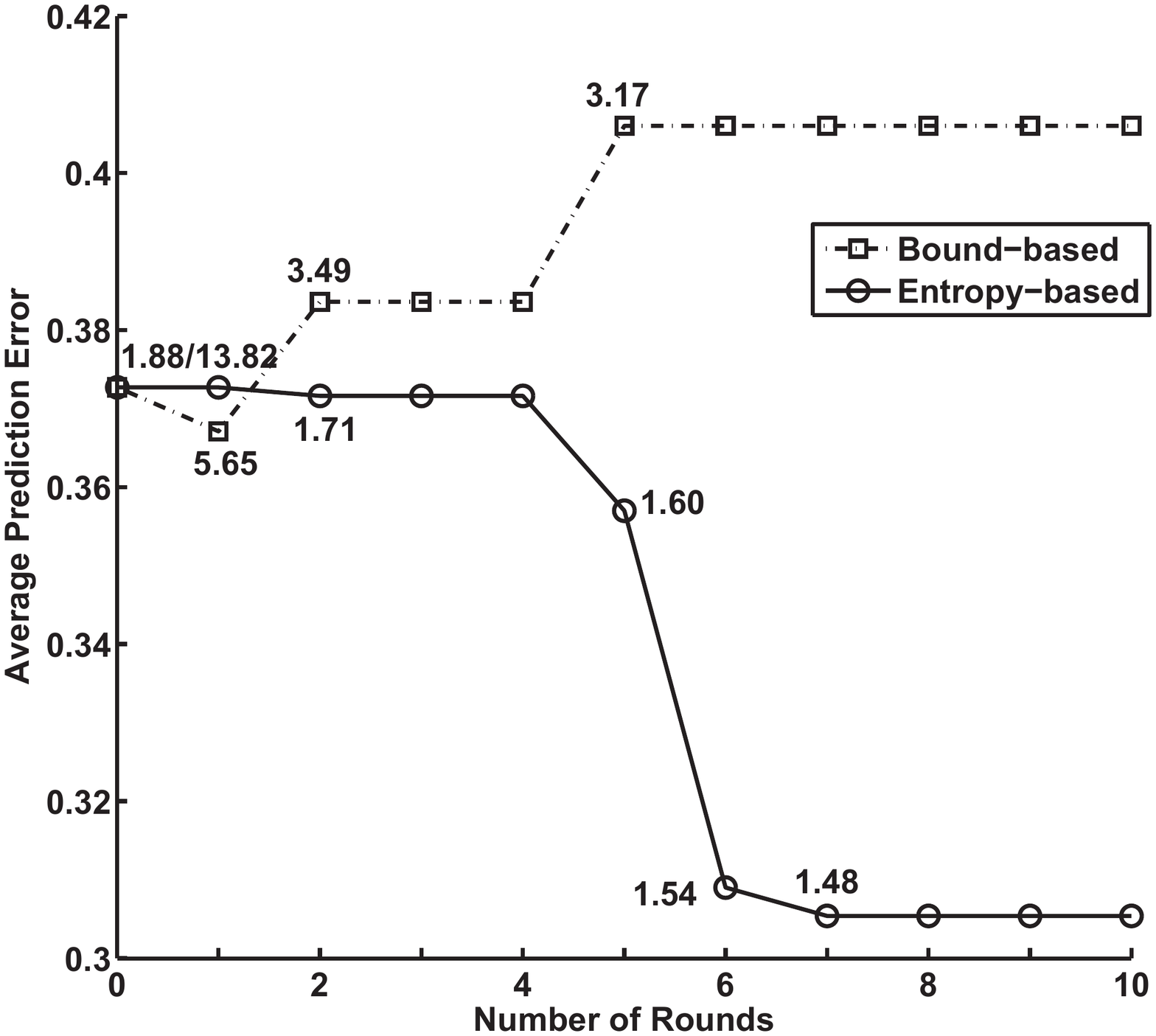}
\centerline{{\scriptsize $(a)$}}
\end{minipage}
\begin{minipage}{3in}
\includegraphics[width=0.8\textwidth]{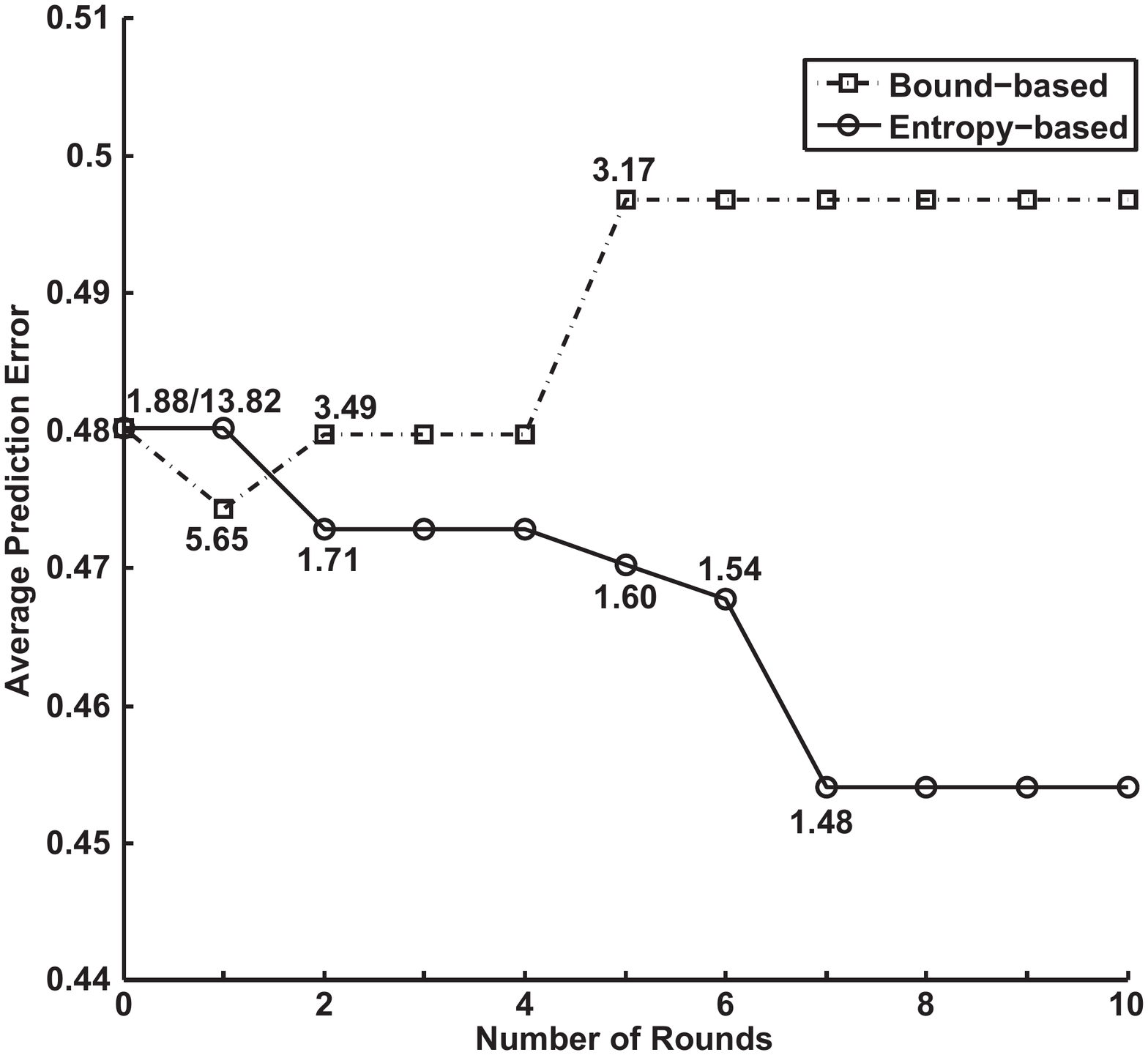}
\centerline{{\scriptsize $(b)$}}
\end{minipage}
\end{minipage}
\caption{{\small ($a$) One-step; ($b$) four-step prediction error of 10 rounds for fix number of tests}}
\label{fig:fix}
\end{figure*}

\begin{figure*}[!ht]
\begin{minipage}{6.0in}
\begin{minipage}{3in}
\includegraphics[width=0.8\textwidth]{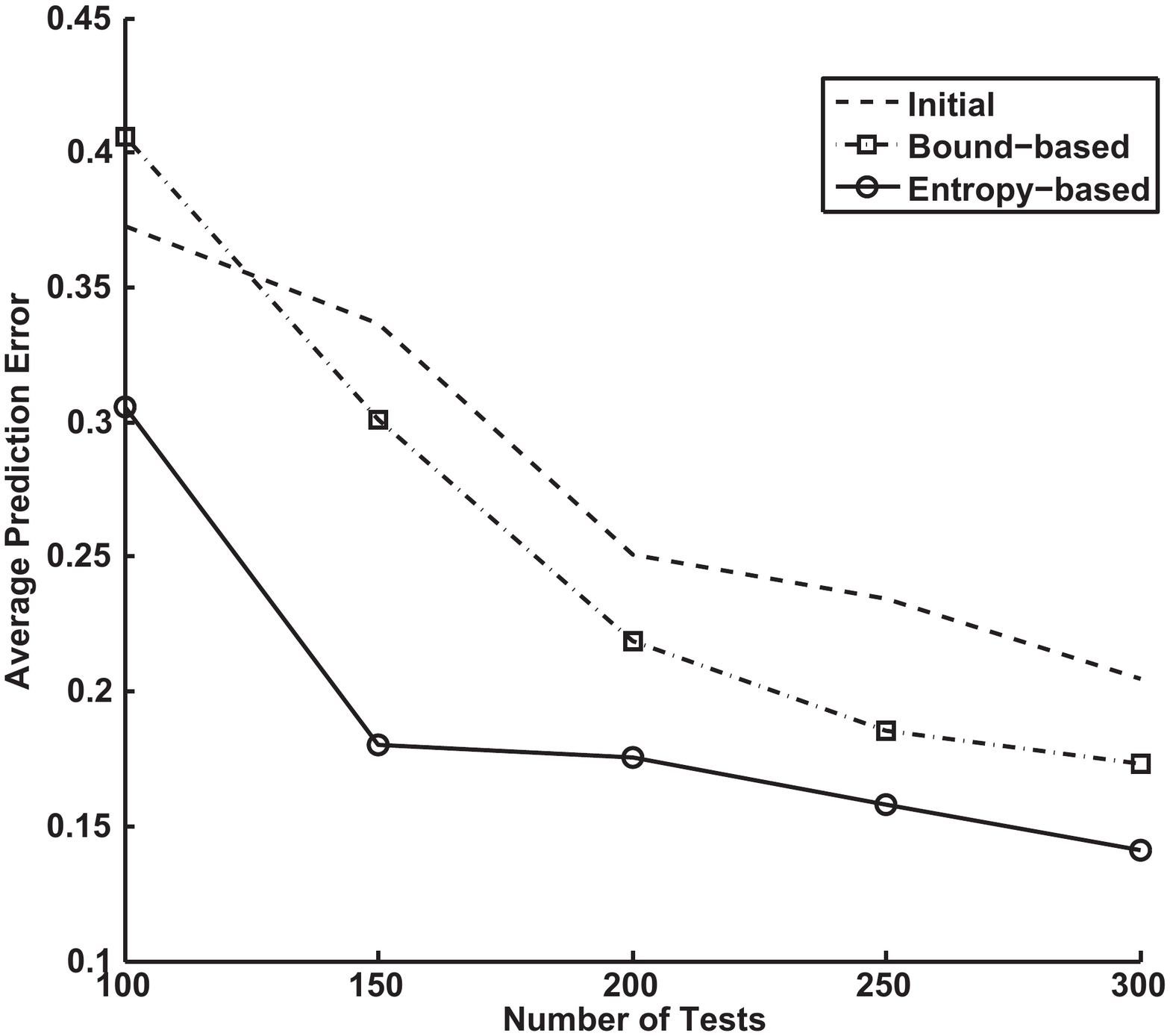}
\centerline{{\scriptsize $(a)$}}
\end{minipage}
\begin{minipage}{3in}
\includegraphics[width=0.8\textwidth]{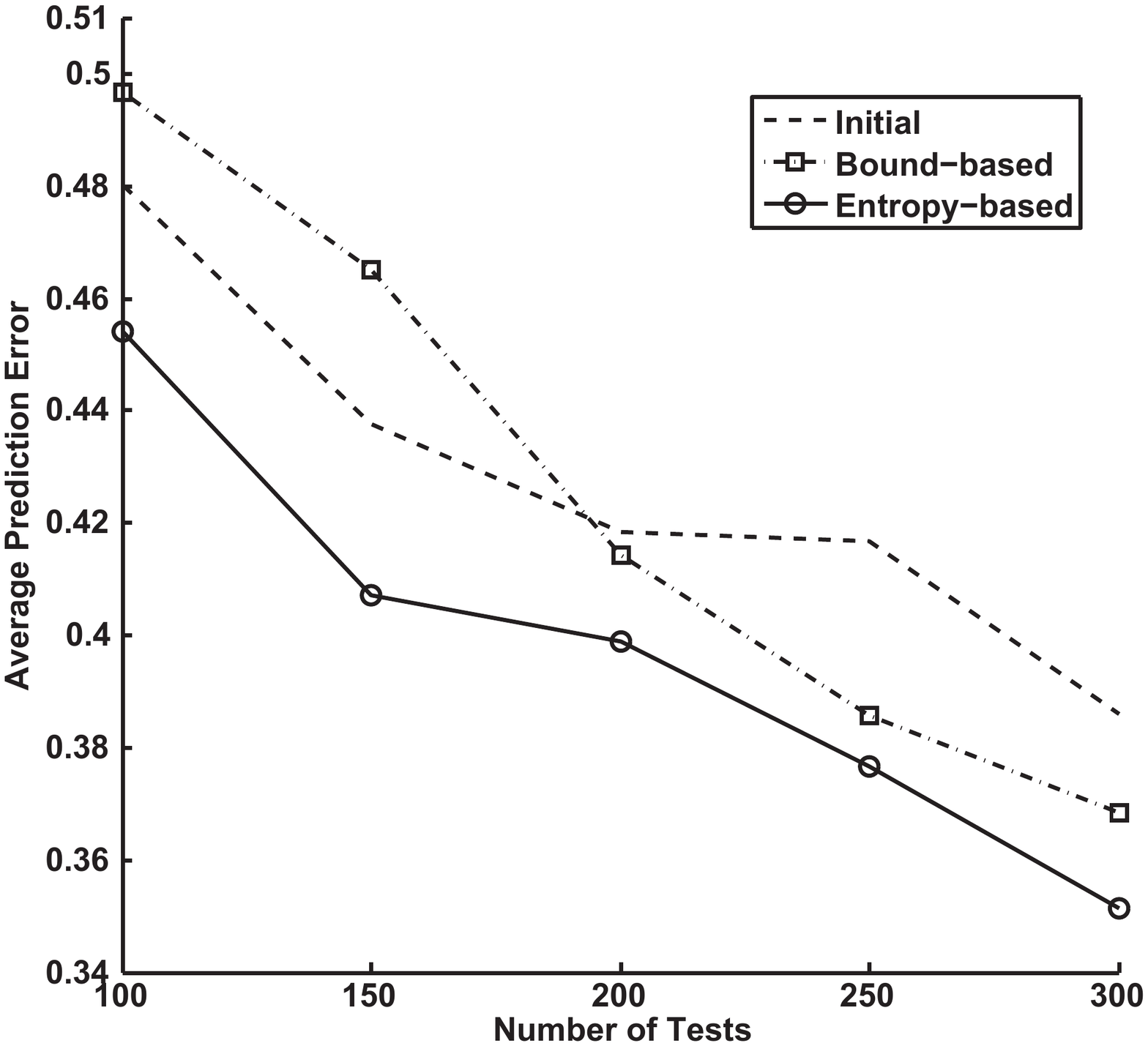}
\centerline{{\scriptsize $(b)$}}
\end{minipage}
\end{minipage}
\caption{{\small ($a$) One-step; ($b$) four-step prediction error of different number of tests}}
\label{fig:vary}
\end{figure*}

Two kinds of experiments were conducted. The first experiment is to evaluate the prediction performance by fixing the number of tests as 100, and ran Algorithm~\ref{Algo:selecting} 10 rounds. For each round, we reported the average one and four prediction errors for both approaches, the results are shown in Fig.~\ref{fig:fix}.  The number at each point in Fig.~\ref{fig:fix} is the average entropy~(for our approach) or the average largest singular values~(for the bound-based approach) in the corresponding round, which also shows that for the entropy-based approach, a higher entropy value results in a lower prediction accuracy while there are no relevance between the singular values and the prediction accuracy. The second experiment is to evaluate the prediction performance by varying the number of tests. For each number of tests, both approaches were ran 10 rounds, and the final results for both one and four-step errors after 10 rounds were reported in Fig.~\ref{fig:vary}. For this experiment, we also reported the initial results~(Initial) without the replacement of the tests.

As can be seen from Figs.~\ref{fig:fix} and~\ref{fig:vary}, both for the one-step and four-step predictions, and both for the two kinds of experiments, in  all the cases, our algorithm performs very well and outperforms the bound-based approaches. Considering the error reported is only the prediction error for one time step, the improvement on prediction accuracy on a long sequence is remarkable. Meanwhile, for the first kind of experiment, as the number of rounds increases, our algorithm reduces its prediction error while the bound-based approaches with increasing rounds do not improve their performances and are very unstable. What also can be seen from the experimental results is that with the increase of the number of tests, for all the approaches, the prediction accuracy of the obtained PSR model also increases, which demonstrates the importance of including more tests in the learning of the models.

\section{Conclusion}

How to choose the bases is a very important problem in spectral learning of PSRs. However, until now, there are very little work that can address this issue successfully. In this paper, by introducing the model entropy for measuring the model accuracy and showing the close relevance between the entropy and model accuracy, we propose an entropy-based basis selection strategy for spectral learning of PSRs. Several experiments were conducted on the PocMan environment, and the results show that compared to the state-of-the-art bound-based approach, our technique is more stable and achieved much better performance.

\acks{This work was supported by the National Natural Science Foundation of China (No. 61375077).}

%\bibliographystyle{plain}
%\bibliography{acml16}

\iffalse
\appendix

\section{First Appendix}\label{apd:first}

This is the first appendix.

\section{Second Appendix}\label{apd:second}

This is the second appendix.
\section
\fi

\end{document}